\documentclass[10pt,twocolumn,letterpaper]{article}

\usepackage{iccv}
\usepackage{times}
\usepackage{epsfig}
\usepackage{graphicx}
\usepackage{amsmath}
\usepackage{amssymb}

\usepackage[thmmarks]{ntheorem}
\usepackage{cases}
{
\theoremstyle{nonumberplain}
\theoremheaderfont{\bfseries}
\theorembodyfont{\normalfont}
\theoremsymbol{\mbox{$\Box$}}

}

\usepackage[breaklinks=true,bookmarks=false]{hyperref}

\iccvfinalcopy 

\def\iccvPaperID{4878} 

\begin{document}

\title{Companion Surface of Danger Cylinder and its Role in Solution Variation of P3P Problem}

\author{Bo Wang\\
Institute of Automation, Chinese Academy of Sciences \\
{\tt\small bo.wang@ia.ac.cn}
\and
Hao Hu \\
Ming Hsieh Department of Electrical Engineering, University of Southern California\\
{\tt\small huhao@usc.edu}
\and
Caixia Zhang \\
College of Science, North China University of Technology\\
{\tt\small zhangcx@ncut.edu.cn}
}

\maketitle

\begin{abstract}
   Traditionally the danger cylinder is intimately related to the solution stability in P3P problem. In this work, we show that the danger cylinder is also closely related to the multiple-solution phenomenon. More specifically, we show when the optical center lies on the danger cylinder, of the 3 possible P3P solutions, i.e., one double solution, and two other solutions, the optical center of the double solution still lies on the danger cylinder, but the optical centers of the other two solutions no longer lie on the danger cylinder. And when the optical center moves on the danger cylinder, accordingly the optical centers of the two other solutions of the corresponding P3P problem form a new surface, characterized by a polynomial equation of degree 12 in the optical center coordinates, called the Companion Surface of Danger Cylinder (CSDC). That means the danger cylinder always has a companion surface. For the significance of CSDC, we show that when the optical center passes through the CSDC, the number of solutions of P3P problem must change by 2. That means CSDC acts as a delimitating surface of the P3P solution space. These new findings shed some new lights on the P3P multi-solution phenomenon, an important issue in PnP study.
\end{abstract}

\section{Introduction}

The Perspective-Three-Point (P3P) problem, the minimal setup of the PnP problem, is to determine the external parameters of a calibrated pinhole camera, based on the correspondence between the three known 3D spatial points and their projected 2D image points. It is a general pose estimation method based on the monocular vision. It was firstly introduced by Grunert in 1841\cite{Grunert1841}. In 1981, Fishler and Bolles\cite{RANSAC1981}introduced the well-known robustness estimation algorithm--RANSAC to the PnP problem, then the P3P problem is popularized in computer vision. The P3P problem requires the minimal setup for camera pose determination, which is especially suitable for restricted environments such as aeronautics or robust estimation problems for discarding outliers or large-scale continuous positioning \cite{high-precision2006}\cite{Kneip2011}\cite{LI2011}\cite{Linnainmaa1988}\cite{Lowe1991}\cite{Nister2007}\cite{Quan1999}\cite{wuyihong2006}. However, the solution to the P3P problem is usually not unique, up to 4 ones \cite{RANSAC1981}. Therefore, before the specific solution to the P3P problem is estimated, the number of solutions should be known in advance to guide the subsequent solution searching.

The related works can be divided into two kinds of ways on the number of solutions of the P3P problem. One is from the algebraic view. Haralick \cite{Haralick1994} summarized 6 different transformation methods. In 2003, Gao et al. \cite{Gaoxiaoshan2003} used Wu's elimination method to triangulate the constrained equations of the P3P problem, and deduced all of the algebraic conditions on the different number of solutions, thus the problem is closed from the algebraic ways.

The other is from the geometric view. Obviously, the geometric conditions are more intuitive. In 1966, Thompson et al. \cite{cylinder1966} connected the singular Jacobian determinant corresponding to the P3P problem with the so-called "danger cylinder", and concluded that the algorithm failed when the optical center lies on the cylinder due to the singularity of the associated Jacobian matrix. In 1991, Wolfe et al. \cite{Wolfe1991} found a condition about four solutions, that is, the optical center lies on a special line perpendicular to the plane of the three control points, and numerical experiments show that the P3P problem has usually two solutions. In 2005, Zhang et al.\cite{four2005} proved that when the optical center lies on one of the three special planes perpendicular to the plane of the three control points, there are also four solutions which constitute two special pairs: one pair is a side-shared solutions, and the other pair is point-shared solutions. In 2006, Zhang et al. \cite{cylinder2006} further showed that when the optical center lies on the danger cylinder, there are usually three solutions: one double solution and two other solutions. In 2010, Sun et al. \cite{sunfengmei2010} found that when the optical center lies on the intersections of the above vertical planes and the danger cylinder, either a pair of side-shared solutions or point-shared solutions degenerates into a double solution, thus the number of solutions decreases from 4 to 3. Since 2012, Reick\cite{Rieck2012JMIV}\cite{Rieck2012VISAPP}\cite{Rieck2014}\cite{Rieck2018}has provided some systematic results via a new algebraic entity on the coordinates of the optical center, which explicitly contains a part closely related to the danger cylinder. In addition, he proved that the necessary and sufficient condition for the P3P problem to have repeated solution is its optical center lying on the danger cylinder.

In this work, we show that when the optical center lies on the danger cylinder, of the 3 possible P3P solutions, the optical center of the double solution still lies on the danger cylinder, but the optical centers of the other two solutions no longer lie on the danger cylinder. In particular, we prove that when the optical center moves on the danger cylinder, the optical centers of the two other solutions of the corresponding P3P problem must lie on a 12-order polynomial surface in the optical center coordinates, called the Companion Surface of Danger Cylinder (CSDC). In addition, we show that when the optical center passes through the CSDC, the number of solutions of P3P problem must change by 2. In other words, the danger cylinder and its companion surface play some critical roles in solution space partitioning.

The paper is organized as follows: Section 2 introduces some preliminaries and new definitions; Section 3 is the main results which include the derivation of the companion surface of danger cylinder and its role in the P3P solution changes; Section 4 concludes the work.

\section{Preliminaries}

For the notational convenience and a better understanding of our main results in the next section, some preliminaries are at first listed in this section.

\subsection{P3P problem and its basic constraint system}

As shown in Fig.1, $A, B, C$ are the three control points with known distances:  $a=|BC|, b=|AC|, c=|AB|$ . $O$ is the optical center of a calibrated camera under the pinhole model. Since the camera is assumed calibrated, the 3 subtended angles $\alpha, \beta, \gamma$  of the projection rays can be computed, hence they can be considered as known entities. Then by the Law of Cosines, the following basic constraint system (\ref{equ1}) can be obtained on the three unknowns: $s_1=|OA|, s_2=|OB|, s_3=|OC|$ .

\begin{gather}\label{equ1}
  \begin{cases}
    s_1^2 +s_2^2 -2cos\gamma s_1s_2 =c^2 \\
    s_1^2 +s_3^2 -2cos \beta s_1s_3 =b^2  \\
    s_2^2 +s_3^2 -2cos\alpha s_2s_3 =a^2
  \end{cases}
\end{gather}

\begin{figure}[t]
\begin{center}
\includegraphics[width=0.5\linewidth]{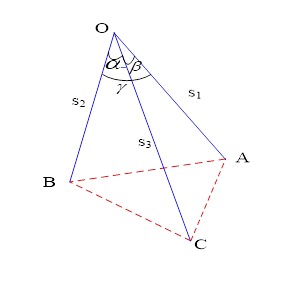}
\end{center}
   \caption{P3P problem definition}
\label{Fig1}
\end{figure}

Note that a triplet satisfying the basic constraint system (\ref{equ1})  does not necessarily mean it is a solution of the P3P problem because such a triplet could contain non-positive elements, or complex elements. For the convenience of subsequent discussions, here we give the following 3 definitions:

{\bf A Constraint-System Solution}: A triplet which satisfies the basic constraint system (\ref{equ1}) .

{\bf A P3P Solution}: A triplet which satisfies the basic constraint system (\ref{equ1})  and all its 3 elements are positive values;

{\bf A Non-P3P Solution}: A triplet which satisfies the basic constraint system (\ref{equ1})  but at least one of its elements is zero, negative or complex value.

Clearly a constraint-system solution is either a P3P solution, or a Non-P3P solution.

Note that if $(s_1,s_2,s_3 )$ is a Constraint-System Solution, -$(s_1,s_2, s_3 )$ must also be a Constraint-System Solution. Following the convention in the P3P literature, $(s_1,s_2,s_3 )$ and -$(s_1,s_2, s_3 )$ are also considered as the same P3P solution in this work.

\subsection{The Danger Cylinder}

As shown in Fig. 2, the vertical cylinder passing through the 3 control points: $A,B,C,$ is called the danger cylinder. It is shown that if the optical center lies on the danger cylinder, the solution of the P3P problem is unstable \cite{cylinder1966}. Later on, Sun et al. \cite{sunfengmei2010} showed that the necessary and sufficient condition for the optical center lying on the danger cylinder is the nullness of the determinant of the Jacobian of the 3 constraints in (\ref{equ1}). In this paper, we will show that the companion surface of danger cylinder also plays a critical role in partitioning the solution space in P3P problem.

Without loss of generality, we assume the 3 control points lie on a unit-circle centered at the origin, hence the danger cylinder can be expressed as:
\begin{gather}\label{unitcircle}
   x^2+y^2-1=0
\end{gather}

\begin{figure}[t]
\begin{center}
\includegraphics[width=0.5\linewidth]{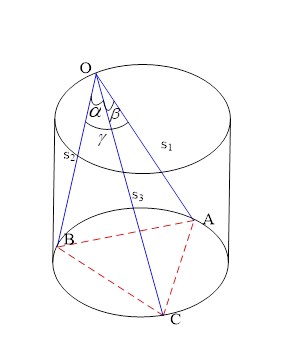}
\end{center}
   \caption{Danger Cylinder}
\label{Fig2}
\end{figure}
\subsection{Companion Surface of Danger Cylinder (CSDC)}

As shown later in this work, when the optical center lies on the danger cylinder, the optical centers of the other two solutions of the corresponding P3P problem must lie on another surface. We call this surface the companion surface of danger cylinder (CSDC). In the next section, we will prove that this CSDC is in fact a polynomial equation of degree 12 in the variables of the optical center's 3 coordinates. Then, we will show that when the optical center passes through the CSDC, the number of P3P solutions always changes by 2: either two non-P3P solutions become two P3P solutions, or two P3P solutions becomes two Non-P3P solutions, depending on the passing direction of the optical center through the CSDC. A simulated CSDC is shown in Fig.3.

\subsection{Rieck's Theorem}

\begin{figure}[t]
\begin{center}
\includegraphics[width=0.5\linewidth]{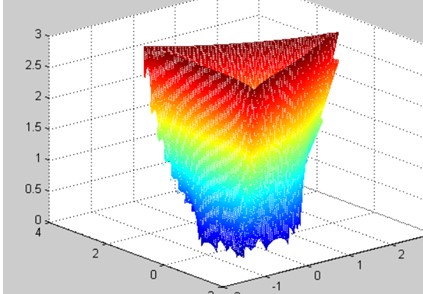}
\end{center}
   \caption{The companion surface of danger cylinder}
\label{Fig3}
\end{figure}
Our result is based on the Rieck's result \cite{Rieck2014}. For the convenience of proof, Rieck's result is here recalled at first.
As in \cite{Rieck2014}, the 3 control points $A,B,C$ and optical center$O$ are defined as:
$$
A=\left [ \begin{array}{c}
    cos\phi_A \\
    sin\phi_A \\
    0
  \end{array}\right ] =\left [ \begin{array}{c}
    x_A \\
    y_A \\
    0
  \end{array}\right ],
B=\left [ \begin{array}{c}
    cos\phi_B \\
    sin\phi_B \\
    0
  \end{array}\right ] =\left [ \begin{array}{c}
    x_B \\
    y_B \\
    0
  \end{array}\right ],
$$

$$
C=\left [ \begin{array}{c}
    cos\phi_C \\
    sin\phi_C \\
    0
  \end{array}\right ] =\left [ \begin{array}{c}
    x_C \\
    y_C \\
    0
  \end{array}\right ],
O=\left [ \begin{array}{c}
    x \\
    y \\
    z
  \end{array}\right ]
$$

Since the direction of the positive $X$-axis can be chosen freely, without loss of generality, we assume $\phi_A+\phi_B+\phi_C=0$  by appropriately choosing the $X$-axis.
Following the Rieck's definitions of the following entities:
$$
\eta = \sqrt{1-cos\phi_A^2-cos\phi_B^2-cos\phi_C^2+2cos\phi_Acos\phi_Bcos\phi_C}
$$
{\small
\begin{gather}\label{eq3}
\begin{split}
&A_A=csc\left( \frac{\phi_B-\phi_C}{2} \right)\\
&\left( sin\frac{\phi_A}{2}\left((x+x_A)^2-(y+y_A)^2 \right) +2cos\frac{\phi_A}{2}(x+x_A)(y+y_A)\right)\\
&B_A=\frac{b^2-c^2}{4}-csc\left( \frac{\phi_B-\phi_C}{2} \right)\\
&\left( sin\frac{\phi_A}{2}\left(\left(x-\frac{x_B+x_C}{2}\right)^2-\left(y-\frac{y_B+y_C}{2}\right)^2 \right)\right.\\
&+\left.2cos\frac{\phi_A}{2}\left(x-\frac{x_B+x_C}{2}\right)\left(y-\frac{y_B+y_C}{2}\right) \right)
\end{split}
\end{gather}
}
{\small
\begin{gather}\label{eq4}
\begin{split}
&A_B=csc\left( \frac{\phi_C-\phi_A}{2} \right)\\
&\left( sin\frac{\phi_B}{2}\left((x+x_B)^2-(y+y_B)^2 \right) +2cos\frac{\phi_B}{2}(x+x_B)(y+y_B)\right)\\
&B_B=\frac{c^2-a^2}{4}-csc\left( \frac{\phi_C-\phi_A}{2} \right)\\
&\left( sin\frac{\phi_B}{2}\left(\left(x-\frac{x_C+x_A}{2}\right)^2-\left(y-\frac{y_C+y_A}{2}\right)^2 \right)\right.\\
&+\left.2cos\frac{\phi_B}{2}\left(x-\frac{x_C+x_A}{2}\right)\left(y-\frac{y_C+y_A}{2}\right) \right)
\end{split}
\end{gather}
}
{\small
\begin{gather}\label{eq5}
\begin{split}
&A_C=csc\left( \frac{\phi_A-\phi_B}{2} \right)\\
&\left( sin\frac{\phi_C}{2}\left((x+x_C)^2-(y+y_C)^2 \right) +2cos\frac{\phi_C}{2}(x+x_C)(y+y_C)\right)\\
&B_C=\frac{a^2-b^2}{4}-csc\left( \frac{\phi_A-\phi_B}{2} \right)\\
&\left( sin\frac{\phi_C}{2}\left(\left(x-\frac{x_A+x_B}{2}\right)^2-\left(y-\frac{y_A+y_B}{2}\right)^2 \right)\right.\\
&+\left.2cos\frac{\phi_C}{2}\left(x-\frac{x_A+x_B}{2}\right)\left(y-\frac{y_A+y_B}{2}\right) \right)
\end{split}
\end{gather}
}
{\small
\begin{gather}\label{eq6}
F_A=\frac{b^2sin\phi_C^2-c^2sin\phi_B^2}{\eta^2}
\end{gather}
}
{\small
\begin{gather}\label{eq7}
F_B=\frac{c^2sin\phi_A^2-a^2sin\phi_C^2}{\eta^2}
\end{gather}
}
{\small
\begin{gather}\label{eq8}
F_C=\frac{a^2sin\phi_B^2-b^2sin\phi_A^2}{\eta^2}
\end{gather}
}

Rieck proved the following equalities by Theorem 1 in \cite{Rieck2014}:

{\small
\begin{gather}
F_A=A_A+B_A\frac{1-x^2-y^2}{z^2}=A_A+C_A
\end{gather}
}
{\small
\begin{gather}\label{eq10}
F_B=A_B+B_B\frac{1-x^2-y^2}{z^2}=A_B+C_B
\end{gather}
}
{\small
\begin{gather}\label{eq11}
F_C=A_C+B_C\frac{1-x^2-y^2}{z^2}=A_C+C_C
\end{gather}
}
In the next section, based on the above equalities, we will derive the constraint equation of CSDC.
\section{Main results}

In section 3.1, we first derive the constraint equation of CSDC. Then, in section 3.2, we prove the proposition that when the optical center passes through the CSDC, the number of P3P solutions always changes by 2.

\subsection{Derivation of constraint equation of CSDC}

{\bf Proposition 1

Given 3 control points, the companion surface of the danger cylinder is a 12-order polynomial in the 3 variables of the optical center coordinates $(x, y, z)$.}

  First, for convenience, we introduce two concepts: F-Property and P-Constraint:

{\bf F-Property}
Note that in (\ref{eq6}), (\ref{eq7}), (\ref{eq8}), $ F_A, F_B, F_C$  are only determined by the coordinates of the 3 control points $A, B, C$ and 3 subtended angles $\alpha, \beta, \gamma$  and they are not explicitly related to the optical center position $(x,y,z)$, which means these three equations hold for all the optical center positions of the P3P solutions. We call this property of (\ref{eq6}), (\ref{eq7}), (\ref{eq8}) F-Property in the next for discussion convenience.

{\bf P-Constraint}
From F-Property, we can derive a polynomial constraint on the optical center coordinates $(x,y,z)$, or when the optical center $O$ lies on either the danger cylinder or the companion surface of danger cylinder (CSDC). This constraint is called P-Constraint.

{\bf Factorization of P-Constraint into danger cylinder and CSDC}

Since P-Constraint is valid for both the optical centers lying on the danger cylinder and CSDC, the P-Constraint can be factorized into the factor $(x^2+y^2-1)$ and another factor, which corresponds to CSDC. Furthermore, we need to determine the degree of the common factor $(x^2+y^2-1)$ in the P-Constraint. Hence our proof mainly consists of two steps: P-Constraint derivation, and factorization of the common factor $(x^2+y^2-1)$ from the P-constraint.

\subsubsection{P-Constraint derivation}

First, when the optical center  $O_0$ lies on the danger cylinder, its corresponding $ F_A,  F_B,  F_C,  A_A,  A_B,  A_C, B_A ,  B_B, B_C$ are denoted as: $ F_{A0},  F_{B0},  F_{C0},  A_{A0},  A_{B0},  A_{C0}, B_{A0} ,  B_{B0}, B_{C0}$; Similarly assume $O_1$ is the optical center of another P3P solution, its corresponding $ F_A,  F_B,  F_C,  A_A,  A_B,  A_C, B_A ,  B_B, B_C$  are denoted as:$ F_{A1},  F_{B1},  F_{C1},  A_{A1},  A_{B1},  A_{C1}, B_{A1} ,  B_{B1}, B_{C1}$ .

Similarly as in \cite{Rieck2014}, Equations (\ref{eq3}) (\ref{eq4}) (\ref{eq5}) can are re-expressed as:
{\small
\begin{gather}
\begin{split}
&A_A=csc\left( \frac{\phi_B-\phi_C}{2} \right)\\
&\left( sin\frac{\phi_A}{2}\left(y^2-x^2+2x-1 \right) +2cos\frac{\phi_A}{2}(xy+y)\right.\\
&\left.+sin\frac{\phi_A}{2}+sin\frac{\phi_A}{2}(y_A^2-x_A^2)+2cos\frac{\phi_A}{2}x_Ay_A \right)\\
&=k_{11}\left(\frac{x^2-y^2-2x+1}{2}\right)+k_{12}(xy+y)+k_{13}
\end{split}
\end{gather}
}
{\small
\begin{gather}\label{eq13}
\begin{split}
&A_B=k_{21}\left(\frac{x^2-y^2-2x+1}{2}\right)+k_{22}(xy+y)+k_{23}
\end{split}
\end{gather}
}
{\small
\begin{gather}\label{eq14}
\begin{split}
&A_C=k_{31}\left(\frac{x^2-y^2-2x+1}{2}\right)+k_{32}(xy+y)+k_{33}
\end{split}
\end{gather}
}
where  $k_{ij}(i,j=1,2,3)$ depend only on the 3 known control points. For our subsequent derivations, their explicit expressions are not needed, hence their explicit forms are not provided here.

Define{\small
\begin{gather*}
\begin{split}
&E_{A1}=\frac{x^2-y^2-2x+1}{2}\\
&E_{A2}=xy+y
\end{split}
\end{gather*}
}
Then, by linear algebraic operation, from (\ref{eq13}) and (\ref{eq14}), $E_{A1}$ and $E_{A2}$ can be re-expressed into:
{\small
\begin{gather}
E_{A1}=E_{11}A_{B}+E_{12}A_{C}+E_{13} \label{eq15}\\
E_{A2}=E_{21}A_{B}+E_{22}A_{C}+E_{23} \label{eq16}
\end{gather}
}
Once again, $E_{ij}  (i=1,2;j=1,2,3)$  are entities, related only to the 3 control points. Assume the optical center $O_0=(x_0,y_0,z_0)$ lies on the danger cylinder in (2). From (\ref{unitcircle}), (\ref{eq15}), (\ref{eq16}) and the definition of $E_{A1}$ and $E_{A2}$ , the following polynomial constraint on $A_{B0}$ ,$A_{C0}$  can be derived:
{\small
\begin{gather}\label{Eq17}
\begin{split}
&(4(E_{11}A_{B0}+E_{12}A_{C0}+E_{13})-1)^3\\
&\left(2(E_{11}A_{B0}+E_{12}A_{C0}+E_{13})^2+2(E_{21}A_{B0}+E_{22}A_{C0}+E_{23})^2\right.\\
&\left.-10(E_{11}A_{B0}+E_{12}A_{C0}+E_{13})-1\right)^2=0
\end{split}
\end{gather}
}

Eq.(\ref{Eq17}) is a polynomial constraint with variables $A_{B0}$ ,$A_{C0}$. As shown in (\ref{eq4}) and (\ref{eq5}), since  $A_{B0}$ ,$A_{C0}$ are only related to the coordinates of the optical center $O_0$, hence Eq. (\ref{Eq17}) is in fact a constraint on the optical center $O_0=(x_0,y_0,z_0)$.

Since $O_0$ lies on the danger cylinder (\ref{unitcircle}), $x_0^2+y_0^2-1=0$, (\ref{eq10}) and (\ref{eq11}) become:
{\small
\begin{gather}
F_{B0}=A_{B0}\\
F_{C0}=A_{C0}
\end{gather}
}
So $F_{B0},F_{C0}$ satisfy Eq.(\ref{Eq17}). According to the F-property,$F_{B1},F_{C1}$ also satisfies Eq.(\ref{Eq17}). Hence Eq.(\ref{Eq20}) holds for $F_{B},F_{C}$ when $(F_B,F_C )=(F_{B1},F_{C1} )$ and $(F_B,F_C )=(F_{B0},F_{C0} )$.In other words,  Eq.(\ref{Eq20}) holds for both optical centers $O_0$ and $O_1$.
{\small
\begin{gather}\label{Eq20}
\begin{split}
&(4(E_{11}F_B+E_{12}F_C+E_{13})-1)^3\\
&\left(2(E_{11}F_B+E_{12}F_{C}+E_{13})^2+2(E_{21}F_B+E_{22}F_C+E_{23})^2\right.\\
&\left.-10(E_{11}F_B+E_{12}F_C+E_{13})-1\right)^2=0
\end{split}
\end{gather}
}
By substituting (\ref{eq10}) and (\ref{eq11}) into Eq.(\ref{Eq20}), we have a polynomial constraint (21) with variables $A_{B},A_{C},C_{B}$  and $C_{C}$. By further substituting (\ref{eq4}), (\ref{eq5}) into (21), we obtain a polynomial constraint equation (22) with optical center coordinates's $x, y, z$ as variables. Eq.(22) is just our P-Constraint. The explicit forms of Eq.(21) and Eq.(22) are provided in Appendix due to their complicated expressions.

Note that P-Constraint in Eq.(22) is valid for the optical centers lying on either the danger cylinder or the CSDC. In order to obtain the constraint on CSDC, we need factorize the factor $(x^2+y^2-1)$ from Eq.(22), then the remaining part should be the constraint on CSDC .

\subsubsection{Factorization of $(x^2+y^2-1)$ from P-Constraint}

For the notational convenience, $(x^2+y^2-1)$ is called DC-factor. As shown in (\ref{eq10}) and (\ref{eq11}), $C_{B}$  and $C_{C}$  both already contains a DC-factor, we should focus on the degree 0 part of  $C_{B}$  and $C_{C}$   in Eq.(21) to figure out whether it also contains a DC-factor.

For the expression convenience, we further define:
\setcounter{equation}{22}
{\small
\begin{gather}\label{Eq23}
\begin{split}
&E_{11}F_B+E_{12}F_C+E_{13}\\
&=(E_{11}A_{B}+E_{12}A_{C}+E_{13})+(E_{11}C_B+E_{12}C_C)\\
&:=EA_1+EC_1
\end{split}
\end{gather}
}
{\small
\begin{gather}\label{Eq24}
\begin{split}
&E_{21}F_B+E_{22}F_C+E_{23}\\
&=(E_{21}A_{B}+E_{22}A_{C}+E_{23})+(E_{21}C_B+E_{22}C_C)\\
&:=EA_2+EC_2
\end{split}
\end{gather}
So, the 0-order part of  $C_{B}$  and $C_{C}$  in Eq.(21) can be expressed as:
\begin{gather}\label{Eq25}
  (4EA_1-1)^3+(2EA_1^2+2EA_2^2-10EA_1-1)^2
\end{gather}
Eq.(\ref{Eq25}) can be simplified into:
{\small
\begin{gather}\label{Eq26}
\begin{split}
  &(1-x^2-y^2)^2\\
  &(x^4-8x^3+2x^2y^2+18x^2+24xy^2+y^4+18y^2-27)
  \end{split}
\end{gather}
}
From Eq.(\ref{Eq26}), we know that the 0-order part of  $C_{B}$  and $C_{C}$ in Eq.(21) contains a DC-factor of multiplicity of 2, or a double DC-factor. To further explore whether the overall order of DC-factor in Eq.(21) is 2, we need to look at the 1-order part of $C_{B}$  and $C_{C}$  in Eq. (21) to figure out whether it indeed contains another DC-Factor, besides the DC-factor, already contained by both  $C_{B}$  and $C_{C}$ , from their definition.

The 1-order part of  $C_{B}$  and $C_{C}$ in Eq.(21) can be expressed as:
{\tiny
\begin{gather}\label{Eq27}
\begin{split}
  &(12(4EA_1-1)^2+2(2EA_1^2+2EA_2^2-10EA_1-1))(4EA_1-10)EC_1\\
  &+8(2EA_1^2+2EA_2^2-10EA_1-1)EA_2EC_2
  \end{split}
\end{gather}
}
By defining: $EB_1=\frac{EC_{1}}{\frac{1-x^2-y^2}{z^2}}$ and $EB_2=\frac{EC_{2}}{\frac{1-x^2-y^2}{z^2}}$  , term (27) becomes term (28).
{\tiny
\begin{gather}\label{Eq28}
\begin{split}
  &(12(4EA_1-1)^2+2(2EA_1^2+2EA_2^2-10EA_1-1))(4EA_1-10)EB_1\\
  &+8(2EA_1^2+2EA_2^2-10EA_1-1)EA_2EB_2
  \end{split}
\end{gather}
}
By some algebraic manipulation, it is shown term (28) indeed contains a 1-order of DC-factor. Hence the overall order of DC-factor in Eq.(22) is 2. By removing this double common factor, DC-factor, in Eq.(22), the remaining part is a 12-order polynomial constraint on optical center  $(x, y, z)$, that is, the constraint equation on CSDC. As the formula of CSDC is too long to show here, the detailed formula is provided in our Maple file named by CSDC.mw. 

In [20], Rieck proved that when the $z$-coordinate of the optical center tends to infinity, a 4-order polynomial, in $(x, y, z)$ is obtained. Our result is in full agreement with Rieck's result. This is because, when $z\rightarrow +\infty$ , the equation of CSDC becomes:
$$
x^4-8x^3+2x^2y^2+18x^2+24xy^2+y^4+18y^2-27=0
$$
It is a 4-order polynomial, and exactly the same constraint equation given by Rieck in \cite{Rieck2014}. A plot of this polynomial constraint is shown in Fig.4.
\begin{figure}[t]
\begin{center}
\includegraphics[width=0.3\linewidth]{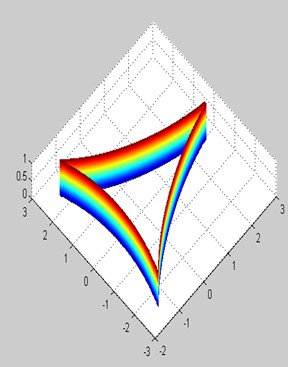}
\includegraphics[width=0.3\linewidth]{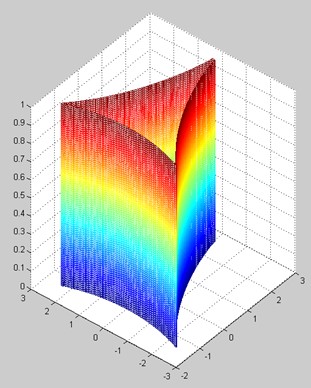}
\end{center}
   \caption{The 4-order toroid when $z\rightarrow +\infty$ }
\label{Fig4}
\end{figure}

\subsection{CSDC is a partitioning surface of P3P solution space}

In section 3.1, we have derived the constraint equation of the CSDC. In this section, we will show that CSDC is a partitioning surface of P3P solution space. More specifically, we show that when the optical center passes through CSDC, the number of P3P solutions always changes by 2: either two Non-P3P solutions become two P3P solutions, or two P3P solutions becomes two Non-P3P solutions, depending on the passing direction of the optical center through the CSDC.

{\bf Proposition 2}

{\bf Given 3 control points $A,B$ and $C$, assume the companion surface of danger cylinder is CSDC$(A, B, C)$ , then when the optical center $O$ passes through CSDC$(A, B, C)$, the number of the P3P solutions always changes by 2.}

Here we would at first point out that: when 2 different triplets of 3 control points lie on a unit circle, they have the same danger cylinder. However, they have different CSDCs. That is, CSDC depends on the 3 control points. That is why we denote our CSDC as CSDC$(A, B, C)$ to explicitly indicate this dependency.

Here is a geometrical interpretation of Proposition 2:

When the optical center passes through CSDC, the elements of a pair of solutions of equation (1) change from real numbers to complex numbers. That means the number of P3P solutions changes.

As shown in Fig.5,  $O_d$ is the optical center, lying on danger cylinder. Its distances to the control points are:$(s_{1d},s_{2d},s_{3d})$  , the subtended angles are: $(\alpha_d,\beta_d,\gamma_d)$. $O_e$is the optical center, lying on CSDC. Its distances to the control points are:$(s_{1e},s_{2e},s_{3e})$   , the subtended angles are: $(\alpha_e,\beta_e,\gamma_e)$. Since $O_d$  and $O_e$  are the two solutions of the same P3P problem, $\alpha_d=\alpha_e,\beta_d,=\beta_e,\gamma_d=\gamma_e$ \footnote {Strictly speaking, the 3 distances of the optical center O to the control points form a P3P solution in Eq.(\ref{equ1}). For simplicity, here we say the optical center $O$ is a solution of Eq.(\ref{equ1})}.

For the convenience of expression, we define:$\phi=cos\alpha, \varphi=cos\beta, \eta = cos\gamma $ . The original P3P constraint system (\ref{equ1}) becomes:
\begin{gather}\label{equ1same}
  \begin{cases}
    s_1^2 +s_2^2 -2 s_1s_2\eta -c^2=0 \\
    s_2^2 +s_3^2 -2 s_2s_3\phi -a^2=0 \\
    s_1^2 +s_3^2 -2 s_1s_3\varphi -b^2=0
  \end{cases}
\end{gather}

\begin{figure}[t]
\begin{center}
\includegraphics[width=0.5\linewidth]{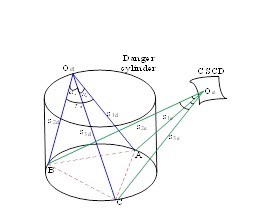}
\end{center}
   \caption{The differential relationship  around $O_d$ and $O_e$  }
\label{Fig5}
\end{figure}

Our proof is mainly about analyzing the differential relationship between $
\left [ \begin{array}{c}
    d\phi \\
    d\varphi \\
    d\eta
  \end{array}\right ]$ and $
\left [ \begin{array}{c}
    ds_1 \\
    ds_2 \\
    ds_3
  \end{array}\right ]$  around  $O_d$  and $O_e$. Our proof is based on the two facts:

(I)	The rank of Jacobian of system (\ref{equ1}) at $O_d$ is 2;

(II)When the optical center $O_e'$ moves on a small enough neighborhood of $O_e$, there is always a point $O_d'$ on a neighborhood of  $O_d$, in complex space, such that the two points $O_e'$ and $O_d'$ are the Constraint-System solutions of the same P3P problem. The fact can be described as: $
\left [ \begin{array}{c}
    d\phi_e \\
    d\varphi_e \\
    d\eta_e
  \end{array}\right ]=\left [ \begin{array}{c}
    d\phi_d \\
    d\varphi_d \\
    d\eta_d
  \end{array}\right ]$ , in differential.

Due to the space limit, here only an outline is provided. The detailed proof is referred to Appendix. Our proof basically consists of the following 3 steps:

{\bf Step 1}

First, from Eq (30), we can get 3 differential approximate constraints on   $
\left [ \begin{array}{c}
    d\phi \\
    d\varphi \\
    d\eta
  \end{array}\right ]$ and $
\left [ \begin{array}{c}
    ds_1 \\
    ds_2 \\
    ds_3
  \end{array}\right ]$.  Then we do the SVD for the Jacobian of equation (30) at $O_d$ , and use its $U, V$ matrices to normalize  $
\left [ \begin{array}{c}
    d\phi \\
    d\varphi \\
    d\eta
  \end{array}\right ]$ and $
\left [ \begin{array}{c}
    ds_1 \\
    ds_2 \\
    ds_3
  \end{array}\right ]$   by:
  $
\left [ \begin{array}{c}
    df_1 \\
    df_2 \\
    df_3
  \end{array}\right ]=U\left [ \begin{array}{c}
    d\phi \\
    d\varphi \\
    d\eta
  \end{array}\right ]$ and $
\left [ \begin{array}{c}
    d\rho_1 \\
    d\rho_2 \\
    d\rho_3
  \end{array}\right ]=V\left [ \begin{array}{c}
    ds_1 \\
    ds_2 \\
    ds_3
  \end{array}\right ]$  . The purpose of this normalization is to ensure that one of the differential constraints does not contain the first order of $d\rho_i$ , according to fact (I);

{\bf Step 2}

We assume the point $O_e'$   approaches to $O_e$  along a straight line. As the Jacobian around $O_e$   is of full rank,  $
\left [ \begin{array}{c}
    d\phi \\
    d\varphi \\
    d\eta
  \end{array}\right ]$   approaches to 0 along a line approximately. The result is that, when $O_e'$    moves through the different side of $O_e$   along a straight line, all the elements of  $
\left [ \begin{array}{c}
    d\phi_e \\
    d\varphi_e \\
    d\eta_e
  \end{array}\right ]$   should change either from negative to positive, or from positive to negative;

{\bf Step 3}

According to fact (II),  $
\left [ \begin{array}{c}
    d\phi_d \\
    d\varphi_d \\
    d\eta_d
  \end{array}\right ]$ also approaches to 0, in the same manner as $
\left [ \begin{array}{c}
    d\phi_e \\
    d\varphi_e \\
    d\eta_e
  \end{array}\right ]$ . By some algebraic manipulation, we prove that to keep fact (II) true, $d\rho_{1d}$, $d\rho_{2d}$ and $d\rho_{3d}^2$  are the infinitesimal quantities of the same order. That further leads to the result that $d\eta$ is approximately proportional to $d\rho_{3d}^2$ .

As in Step 2,  $d\eta$ can be either positive or negative when the point $O_e'$  lies in the different side of the line with respect to $O_e$  . As a result, $d\rho_{3d}^2$  will be either positive or negative, accordingly. So, $d\rho_{3d}$  will be either two real numbers or a pair of complex numbers with non-zero imaginary part. It is shown in Appendix X that when $d\rho_{3d}$   is a pair of real numbers, the corresponding Constraint-System solutions are two P3P solutions, but when it is a pair of complex numbers, the corresponding Constraint-System solutions are two Non-P3P solutions. In sum, the number of P3P solutions always changes by 2.

\section{Conclusion}

In this work, we show that the danger cylinder has a companion surface, which is a 12-order polynomial in the variable of the optical center $(x, y, z)$. In addition, we find the CSDC plays an important role in partitioning the distribution space of the P3P problem solution. More specifically, we find that when the optical center passes through the companion surface, the number of P3P solutions always changes by 2, either 2 Non-P3P solutions become 2 P3P solutions, or 2 P3P solutions become 2 Non-P3P solutions, depending on the passing direction. To our knowledge, we are the first in the literature to discover the existence of a companion surface of the danger cylinder and its surprising role in multi-solution phenomenon in the P3P problem.
{\small
\bibliographystyle{ieee}
\bibliography{egbibmythird}
}

\section*{Appendix (about the proof of section 3.2)}
The 3 steps of the proof in Sections 3.2 are detailed below:\\
{\bf Step 1} Around $\left( {{s_{1d}},{s_{2d}},{s_{3d}}} \right)$, the first order differential relationship between $\left[ {\begin{array}{*{20}{c}}
{d{\phi _d}}\\
{d{\varphi _d}}\\
{d{\eta _d}}
\end{array}} \right]$ and $\left[ {\begin{array}{*{20}{c}}
{d{s_{1d}}}\\
{d{s_{2d}}}\\
{d{s_{3d}}}
\end{array}} \right]$ is:
\begin{gather}\label{eq31}
\begin{array}{l}
\left[ {\begin{array}{*{20}{c}}
{d{\phi _d}}\\
{d{\varphi _d}}\\
{d{\eta _d}}
\end{array}} \right] = \left[ {\begin{array}{*{20}{c}}
{{\raise0.7ex\hbox{${\partial \phi }$} \!\mathord{\left/
 {\vphantom {{\partial \phi } {\partial {s_1}}}}\right.\kern-\nulldelimiterspace}
\!\lower0.7ex\hbox{${\partial {s_1}}$}}}&{{\raise0.7ex\hbox{${\partial \phi }$} \!\mathord{\left/
 {\vphantom {{\partial \phi } {\partial {s_2}}}}\right.\kern-\nulldelimiterspace}
\!\lower0.7ex\hbox{${\partial {s_2}}$}}}&{{\raise0.7ex\hbox{${\partial \phi }$} \!\mathord{\left/
 {\vphantom {{\partial \phi } {\partial {s_3}}}}\right.\kern-\nulldelimiterspace}
\!\lower0.7ex\hbox{${\partial {s_3}}$}}}\\
{{\raise0.7ex\hbox{${\partial \varphi }$} \!\mathord{\left/
 {\vphantom {{\partial \varphi } {\partial {s_1}}}}\right.\kern-\nulldelimiterspace}
\!\lower0.7ex\hbox{${\partial {s_1}}$}}}&{{\raise0.7ex\hbox{${\partial \varphi }$} \!\mathord{\left/
 {\vphantom {{\partial \varphi } {\partial {s_2}}}}\right.\kern-\nulldelimiterspace}
\!\lower0.7ex\hbox{${\partial {s_2}}$}}}&{{\raise0.7ex\hbox{${\partial \varphi }$} \!\mathord{\left/
 {\vphantom {{\partial \varphi } {\partial {s_3}}}}\right.\kern-\nulldelimiterspace}
\!\lower0.7ex\hbox{${\partial {s_3}}$}}}\\
{{\raise0.7ex\hbox{${\partial \eta }$} \!\mathord{\left/
 {\vphantom {{\partial \eta } {\partial {s_1}}}}\right.\kern-\nulldelimiterspace}
\!\lower0.7ex\hbox{${\partial {s_1}}$}}}&{{\raise0.7ex\hbox{${\partial \eta }$} \!\mathord{\left/
 {\vphantom {{\partial \eta } {\partial {s_2}}}}\right.\kern-\nulldelimiterspace}
\!\lower0.7ex\hbox{${\partial {s_2}}$}}}&{{\raise0.7ex\hbox{${\partial \eta }$} \!\mathord{\left/
 {\vphantom {{\partial \eta } {\partial {s_3}}}}\right.\kern-\nulldelimiterspace}
\!\lower0.7ex\hbox{${\partial {s_3}}$}}}
\end{array}} \right]\left[ {\begin{array}{*{20}{c}}
{d{s_{1d}}}\\
{d{s_{2d}}}\\
{d{s_{3d}}}
\end{array}} \right]\\
 \approx \left[ {\begin{array}{*{20}{c}}
0&{\frac{{{s_{21}} - {s_{31}}\phi }}{{{s_{21}}{s_{31}}}}}&{\frac{{{s_{31}} - {s_{21}}\phi }}{{{s_{21}}{s_{31}}}}}\\
{\frac{{{s_{11}} - {s_{31}}\varphi }}{{{s_{31}}{s_{11}}}}}&0&{\frac{{{s_{31}} - {s_{11}}\varphi }}{{{s_{31}}{s_{11}}}}}\\
{\frac{{{s_{11}} - {s_{21}}\eta }}{{{s_{11}}{s_{21}}}}}&{\frac{{{s_{21}} - {s_{11}}\eta }}{{{s_{11}}{s_{21}}}}}&0
\end{array}} \right]\left[ {\begin{array}{*{20}{c}}
{d{s_{1d}}}\\
{d{s_{2d}}}\\
{d{s_{3d}}}
\end{array}} \right]\\
 = {J_{11}}\left[ {\begin{array}{*{20}{c}}
{d{s_{1d}}}\\
{d{s_{2d}}}\\
{d{s_{3d}}}
\end{array}} \right]
\end{array} \end{gather}
 As stated by Sun \cite{sunfengmei2010}, when the optical center lies on the danger cylinder, the rank of $J_{11}$  is 2. So, by SVD decomposition of $J_{11}$ :
\begin{gather}\label{eq32}
{J_{11}} = \left[ {\begin{array}{*{20}{c}}
{{u_{11}}}&{{u_{21}}}&{{u_{31}}}
\end{array}} \right]\left[ {\begin{array}{*{20}{c}}
{{\lambda _{11}}}&{}&{}\\
{}&{{\lambda _{21}}}&{}\\
{}&{}&0
\end{array}} \right]\left[ {\begin{array}{*{20}{c}}
{v_{21}^T}\\
{v_{21}^T}\\
{v_{31}^T}
\end{array}} \right]
\end{gather}
Substituting (\ref{eq32}) to (\ref{eq31}), we have:
\begin{gather}\label{eq33}
\left[ {\begin{array}{*{20}{c}}
{u_{21}^T}\\
{u_{21}^T}\\
{u_{31}^T}
\end{array}} \right]\left[ {\begin{array}{*{20}{c}}
{d{\phi _d}}\\
{d{\varphi _d}}\\
{d{\eta _d}}
\end{array}} \right] = \left[ {\begin{array}{*{20}{c}}
{{\lambda _{11}}}&{}&{}\\
{}&{{\lambda _{21}}}&{}\\
{}&{}&0
\end{array}} \right]\left[ {\begin{array}{*{20}{c}}
{v_{11}^T}\\
{v_{21}^T}\\
{v_{31}^T}
\end{array}} \right]\left[ {\begin{array}{*{20}{c}}
{d{s_{1d}}}\\
{d{s_{2d}}}\\
{d{s_{3d}}}
\end{array}} \right]\begin{array}{*{20}{c}}
{}&{}&{}&{}
\end{array}\end{gather}
From equation (\ref{eq33}), we have the following equation:
$$u_{31}^T\left[ {\begin{array}{*{20}{c}}
{d{\phi _d}}\\
{d{\varphi _d}}\\
{d{\eta _d}}
\end{array}} \right] = 0$$
Then we define new terms $f_1,f_2,f_3$, by:
$$\left[ {\begin{array}{*{20}{c}}
{{f_1}}\\
{{f_2}}\\
{{f_3}}
\end{array}} \right] = \left[ {\begin{array}{*{20}{c}}
{u_{21}^T}\\
{u_{21}^T}\\
{u_{31}^T}
\end{array}} \right]\left[ {\begin{array}{*{20}{c}}
\phi \\
\varphi \\
\eta
\end{array}} \right]$$
By defining:
$\left[ {\begin{array}{*{20}{c}}
{{\rho _1}}\\
{{\rho _2}}\\
{{\rho _3}}
\end{array}} \right] = \left[ {\begin{array}{*{20}{c}}
{v_{11}^T}\\
{v_{21}^T}\\
{v_{31}^T}
\end{array}} \right]\left[ {\begin{array}{*{20}{c}}
{{s_1}}\\
{{s_2}}\\
{{s_3}}
\end{array}} \right]$
we have:
$$\left[ {\begin{array}{*{20}{c}}
{d{\rho _1}}\\
{d{\rho _2}}\\
{d{\rho _3}}
\end{array}} \right] = \left[ {\begin{array}{*{20}{c}}
{v_{11}^T}\\
{v_{21}^T}\\
{v_{31}^T}
\end{array}} \right]\left[ {\begin{array}{*{20}{c}}
{d{s_1}}\\
{d{s_2}}\\
{d{s_3}}
\end{array}} \right]$$
For ${f_i}({i = 1,2,3})$,around $\left( {{s_{1d}},{s_{2d}},{s_{3d}}} \right)$ by making a second order differential approximation, we have:
\begin{gather} \label{eq34}
\begin{array}{l}
\left[ {\begin{array}{*{20}{c}}
{d{f_{1d}}}\\
{d{f_{2d}}}\\
{d{f_{3d}}}
\end{array}} \right]
\approx \left[ {\begin{array}{*{20}{c}}
{\frac{{\partial {f_1}}}{{\partial {\rho _1}}}}&{\frac{{\partial {f_1}}}{{\partial {\rho _2}}}}&{\frac{{\partial {f_1}}}{{\partial {\rho _3}}}}\\
{\frac{{\partial {f_2}}}{{\partial {\rho _1}}}}&{\frac{{\partial {f_2}}}{{\partial {\rho _2}}}}&{\frac{{\partial {f_2}}}{{\partial {\rho _3}}}}\\
{\frac{{\partial {f_3}}}{{\partial {\rho _1}}}}&{\frac{{\partial {f_3}}}{{\partial {\rho _2}}}}&{\frac{{\partial {f_3}}}{{\partial {\rho _3}}}}
\end{array}} \right]\left[ {\begin{array}{*{20}{c}}
{d{\rho _{1d}}}\\
{d{\rho _{2d}}}\\
{d{\rho _{3d}}}
\end{array}} \right]\\
{\rm{          }} + \left[ {\begin{array}{*{20}{c}}
{{{\left[ {\begin{array}{*{20}{c}}
{d{\rho _{1d}}}\\
{d{\rho _{2d}}}\\
{d{\rho _{3d}}}
\end{array}} \right]}^T}{H_{1d}}\left[ {\begin{array}{*{20}{c}}
{d{\rho _{1d}}}\\
{d{\rho _{2d}}}\\
{d{\rho _{3d}}}
\end{array}} \right]}\\
{{{\left[ {\begin{array}{*{20}{c}}
{d{\rho _{1d}}}\\
{d{\rho _{2d}}}\\
{d{\rho _{3d}}}
\end{array}} \right]}^T}{H_{2d}}\left[ {\begin{array}{*{20}{c}}
{d{\rho _{1d}}}\\
{d{\rho _{2d}}}\\
{d{\rho _{3d}}}
\end{array}} \right]}\\
{{{\left[ {\begin{array}{*{20}{c}}
{d{\rho _{1d}}}\\
{d{\rho _{2d}}}\\
{d{\rho _{3d}}}
\end{array}} \right]}^T}{H_{3d}}\left[ {\begin{array}{*{20}{c}}
{d{\rho _{1d}}}\\
{d{\rho _{2d}}}\\
{d{\rho _{3d}}}
\end{array}} \right]}
\end{array}} \right]\\
 = \left[ {\begin{array}{*{20}{c}}
{{\lambda _{11}}d{\rho _{1d}} + {{\left[ {\begin{array}{*{20}{c}}
{d{\rho _{1d}}}\\
{d{\rho _{2d}}}\\
{d{\rho _{3d}}}
\end{array}} \right]}^T}{H_{1d}}\left[ {\begin{array}{*{20}{c}}
{d{\rho _{1d}}}\\
{d{\rho _{2d}}}\\
{d{\rho _{3d}}}
\end{array}} \right]}\\
{{\lambda _{21}}d{\rho _{2d}} + {{\left[ {\begin{array}{*{20}{c}}
{d{\rho _{1d}}}\\
{d{\rho _{2d}}}\\
{d{\rho _{3d}}}
\end{array}} \right]}^T}{H_{2d}}\left[ {\begin{array}{*{20}{c}}
{d{\rho _{1d}}}\\
{d{\rho _{2d}}}\\
{d{\rho _{3d}}}
\end{array}} \right]}\\
{{{\left[ {\begin{array}{*{20}{c}}
{d{\rho _{1d}}}\\
{d{\rho _{2d}}}\\
{d{\rho _{3d}}}
\end{array}} \right]}^T}{H_{3d}}\left[ {\begin{array}{*{20}{c}}
{d{\rho _{1d}}}\\
{d{\rho _{2d}}}\\
{d{\rho _{3d}}}
\end{array}} \right]}
\end{array}} \right]
\end{array}
\end{gather}
Based on equation (\ref{eq34}), we further define:
\begin{gather}\label{eq35}
\begin{array}{l}
\left[ {\begin{array}{*{5}{c}}
{d{h_{1d}}}\\
{d{h_{2d}}}\\
{d{h_{3d}}}
\end{array}} \right] = \left[ {\begin{array}{*{20}{c}}
{d{f_{1d}} + d{f_{2d}}}\\
{d{f_{1d}} - d{f_{2d}}}\\
{d{f_{3d}}}
\end{array}} \right]\\
 \approx \left[ {\begin{array}{*{20}{c}}
{{\lambda _{11}}d{\rho _{1d}} + {\lambda _{21}}d{\rho _{2d}} + {{\left[ {\begin{array}{*{20}{c}}
{d{\rho _{1d}}}\\
{d{\rho _{2d}}}\\
{d{\rho _{3d}}}
\end{array}} \right]}^T}\left( {{H_{1d}} + {H_{2d}}} \right)\left[ {\begin{array}{*{20}{c}}
{d{\rho _{1d}}}\\
{d{\rho _{2d}}}\\
{d{\rho _{3d}}}
\end{array}} \right]}\\
{{\lambda _{11}}d{\rho _{1d}} - {\lambda _{21}}d{\rho _{2d}} + {{\left[ {\begin{array}{*{20}{c}}
{d{\rho _{1d}}}\\
{d{\rho _{2d}}}\\
{d{\rho _{3d}}}
\end{array}} \right]}^T}\left( {{H_{1d}} - {H_{2d}}} \right)\left[ {\begin{array}{*{20}{c}}
{d{\rho _{1d}}}\\
{d{\rho _{2d}}}\\
{d{\rho _{3d}}}
\end{array}} \right]}\\
{\left[ {\begin{array}{*{20}{c}}
{d{\rho _{1d}}}\\
{d{\rho _{2d}}}\\
{d{\rho _{3d}}}
\end{array}} \right]}
\end{array}} \right]\\
 = \left[ {\begin{array}{*{20}{c}}
{{\lambda _{11}}d{\rho _{1d}} + {\lambda _{21}}d{\rho _{2d}} + {{\left[ {\begin{array}{*{20}{c}}
{d{\rho _1}}\\
{d{\rho _2}}\\
{d{\rho _3}}
\end{array}} \right]}^T}{H_{1d}}'\left[ {\begin{array}{*{20}{c}}
{d{\rho _1}}\\
{d{\rho _2}}\\
{d{\rho _3}}
\end{array}} \right]}\\
{{\lambda _{11}}d{\rho _{1d}} - {\lambda _{21}}d{\rho _{2d}} + {{\left[ {\begin{array}{*{20}{c}}
{d{\rho _1}}\\
{d{\rho _2}}\\
{d{\rho _3}}
\end{array}} \right]}^T}{H_{2d}}'\left[ {\begin{array}{*{20}{c}}
{d{\rho _1}}\\
{d{\rho _2}}\\
{d{\rho _3}}
\end{array}} \right]}\\
{\left[ {\begin{array}{*{20}{c}}
{d{\rho _{1d}}}\\
{d{\rho _{2d}}}\\
{d{\rho _{3d}}}
\end{array}} \right]}
\end{array}} \right]
\end{array}
\end{gather}
As shown in Fig.(5), ${O_c}$, with $\left( {{s_{1c}},{s_{2c}},{s_{3c}}} \right)$ , lying on CSDC, is another solution to the P3P problem, corresponding with the solution ${O_d}$. For the optical center, around ${O_c}$, we have the differential constraints:
\begin{gather}\label{eq36}
\begin{array}{l}
\left[ {\begin{array}{*{20}{c}}
{d{h_{1c}}}\\
{d{h_{2c}}}\\
{d{h_{3c}}}
\end{array}} \right] = \left[ {\begin{array}{*{20}{c}}
{d{g_{1c}} + d{g_{2c}}}\\
{d{g_{1c}} - d{g_{2c}}}\\
{d{g_{3c}}}
\end{array}} \right]\\
 \approx J'\left[ {\begin{array}{*{20}{c}}
{d{\rho _{1c}}}\\
{d{\rho _{2c}}}\\
{d{\rho _{3c}}}
\end{array}} \right] + \left[ {\begin{array}{*{20}{c}}
{{{\left[ {\begin{array}{*{20}{c}}
{d{\rho _{1c}}}\\
{d{\rho _{2c}}}\\
{d{\rho _{3c}}}
\end{array}} \right]}^T}{H_{1c}}'\left[ {\begin{array}{*{20}{c}}
{d{\rho _{1c}}}\\
{d{\rho _{2c}}}\\
{d{\rho _{3c}}}
\end{array}} \right]}\\
{{{\left[ {\begin{array}{*{20}{c}}
{d{\rho _{1c}}}\\
{d{\rho _{2c}}}\\
{d{\rho _{3c}}}
\end{array}} \right]}^T}{H_{2c}}'\left[ {\begin{array}{*{20}{c}}
{d{\rho _{1c}}}\\
{d{\rho _{2c}}}\\
{d{\rho _{3c}}}
\end{array}} \right]}\\
{{{\left[ {\begin{array}{*{20}{c}}
{d{\rho _{1c}}}\\
{d{\rho _{2c}}}\\
{d{\rho _{3c}}}
\end{array}} \right]}^T}{H_{3c}}'\left[ {\begin{array}{*{20}{c}}
{d{\rho _{1c}}}\\
{d{\rho _{2c}}}\\
{d{\rho _{3c}}}
\end{array}} \right]}
\end{array}} \right]
\end{array}
\end{gather}
The difference between equation (\ref{eq35}) and (\ref{eq36}), is that $J'$  is a full rank matrix. So, for any full rank linear translation of  $\left[ {\begin{array}{*{20}{c}}
{d{h_{1c}}}\\
{d{h_{2c}}}\\
{d{h_{3c}}}
\end{array}} \right]$, the generated 3 new differential constraints must all contain the first order of $\left[ {\begin{array}{*{20}{c}}
{d{\rho _{1c}}}\\
{d{\rho _{2c}}}\\
{d{\rho _{3c}}}
\end{array}} \right]$ . So, the equation (\ref{eq36}) can ignore the second order of $\left[ {\begin{array}{*{20}{c}}
{d{\rho _{1c}}}\\
{d{\rho _{2c}}}\\
{d{\rho _{3c}}}
\end{array}} \right]$, and we have:
\begin{gather}\label{eq37}
\left[ {\begin{array}{*{20}{c}}
{d{h_{1c}}}\\
{d{h_{2c}}}\\
{d{h_{3c}}}
\end{array}} \right] \approx J'\left[ {\begin{array}{*{20}{c}}
{d{\rho _{1c}}}\\
{d{\rho _{2c}}}\\
{d{\rho _{3c}}}
\end{array}} \right]
\end{gather}

So, we get the fact: at the region nearby$\left( {{s_{1c}},{s_{2c}},{s_{3c}}} \right)$, vector $\left[ {\begin{array}{*{20}{c}}
{d{h_{1c}}}\\
{d{h_{2c}}}\\
{d{h_{3c}}}
\end{array}} \right]$ is linear with vector $\left[ {\begin{array}{*{20}{c}}
{d{\rho _{1c}}}\\
{d{\rho _{2c}}}\\
{d{\rho _{3c}}}
\end{array}} \right]$.

{\bf Step 2}

So, when $\left[ {\begin{array}{*{20}{c}}
{d{\rho _{1c}}}\\
{d{\rho _{2c}}}\\
{d{\rho _{3c}}}
\end{array}} \right]$ approaches ${\left( {0,0,0} \right)^T}$ in a fixed direction, $\left[ {\begin{array}{*{20}{c}}
{d{h_{1c}}}\ \
{d{h_{2c}}}\\
{d{h_{3c}}} \end{array}} \right]$ also approaches ${\left( {0,0,0} \right)^T}$ in fixed direction, defined as $D = \left[ {\begin{array}{*{20}{c}}
{{D_1}}\\
{{D_2}}\\
{{D_3}}
\end{array}} \right]$. So, we have:
\begin{gather}
\left[ {\begin{array}{*{20}{c}}
{d{h_{1c}}}\\
{d{h_{2c}}}\\
{d{h_{3c}}}
\end{array}} \right] \approx k\left[ {\begin{array}{*{20}{c}}
{{D_1}}\\
{{D_2}}\\
{{D_3}}
\end{array}} \right],k \to 0
\end{gather}
As fact (II), when an optical center is nearby $\left( {{s_{1c}},{s_{2c}},{s_{3c}}} \right)$, another solution of the same P3P-constraint equation is nearby $\left( {{s_{1d}},{s_{2d}},{s_{3d}}} \right)$, in the complex space, the imaginary part of which can be either non-zero or zero.
The below equation always holds:
\[\left[ {\begin{array}{*{20}{c}}
{d{h_{1d}}}\\
{d{h_{2d}}}\\
{d{h_{3d}}}
\end{array}} \right] = \left[ {\begin{array}{*{20}{c}}
{d{h_{1c}}}\\
{d{h_{2c}}}\\
{d{h_{3c}}}
\end{array}} \right] \approx k\left[ {\begin{array}{*{20}{c}}
{{D_1}}\\
{{D_2}}\\
{{D_3}}
\end{array}} \right],k \to 0\]
In the condition that $D_1,D_2,D_3$,  do all not equal to 0,we have:
\[\left\{ \begin{array}{l}
\frac{{d{h_{1d}}}}{{d{h_{3d}}}} \approx \frac{{{D_1}}}{{{D_3}}}\\
\frac{{d{h_{2d}}}}{{d{h_{3d}}}} \approx \frac{{{D_2}}}{{{D_3}}}
\end{array} \right.\]

{\bf Step 3}

So, we have:
\begin{gather*}\left\{ \begin{array}{l}
{D_3}\left( {{\lambda _{11}}d{\rho _{1d}} + {\lambda _{21}}d{\rho _{2d}} + {{\left[ {\begin{array}{*{20}{c}}
{d{\rho _{1d}}}\\
{d{\rho _{2d}}}\\
{d{\rho _{3d}}}
\end{array}} \right]}^T}{H_{1d}}'\left[ {\begin{array}{*{20}{c}}
{d{\rho _{1d}}}\\
{d{\rho _{2d}}}\\
{d{\rho _{3d}}}
\end{array}} \right]} \right)\\
 - {D_1}{\left[ {\begin{array}{*{20}{c}}
{d{\rho _{1d}}}\\
{d{\rho _{2d}}}\\
{d{\rho _{3d}}}
\end{array}} \right]^T}{H_{3d}}'\left[ {\begin{array}{*{20}{c}}
{d{\rho _{1d}}}\\
{d{\rho _{2d}}}\\
{d{\rho _{3d}}}
\end{array}} \right] \approx 0\\
{D_3}\left( {{\lambda _{11}}d{\rho _{1d}} - {\lambda _{21}}d{\rho _{2d}} + {{\left[ {\begin{array}{*{20}{c}}
{d{\rho _{1d}}}\\
{d{\rho _{2d}}}\\
{d{\rho _{3d}}}
\end{array}} \right]}^T}{H_{2d}}'\left[ {\begin{array}{*{20}{c}}
{d{\rho _{1d}}}\\
{d{\rho _{2d}}}\\
{d{\rho _{3d}}}
\end{array}} \right]} \right)\\
 - {D_2}{\left[ {\begin{array}{*{20}{c}}
{d{\rho _{1d}}}\\
{d{\rho _{2d}}}\\
{d{\rho _{3d}}}
\end{array}} \right]^T}{H_{3d}}'\left[ {\begin{array}{*{20}{c}}
{d{\rho _{1d}}}\\
{d{\rho _{2d}}}\\
{d{\rho _{3d}}}
\end{array}} \right] \approx 0
\end{array} \right.\end{gather*}

\begin{numcases}{}
{D_3}{\lambda _{11}}d{\rho _{1d}} + {D_3}{\lambda _{21}}d{\rho _{2d}} \notag \\ \label{eq39}
+ {\left[ {\begin{array}{*{20}{c}}
{d{\rho _{1d}}}\\
{d{\rho _{2d}}}\\
{d{\rho _{3d}}}
\end{array}} \right]^T}{H_{4d}}'\left[ {\begin{array}{*{20}{c}}
{d{\rho _{1d}}}\\
{d{\rho _{2d}}}\\
{d{\rho _{3d}}}
\end{array}} \right] \approx 0 \\
{D_3}{\lambda _{11}}d{\rho _{1d}} - {D_3}{\lambda _{21}}d{\rho _{2d}} \notag\\ \label{eq40}
+ {\left[ {\begin{array}{*{20}{c}}
{d{\rho _{1d}}}\\
{d{\rho _{2d}}}\\
{d{\rho _{3d}}}
\end{array}} \right]^T}{H_{5d}}'\left[ {\begin{array}{*{20}{c}}
{d{\rho _{1d}}}\\
{d{\rho _{2d}}}\\
{d{\rho _{3d}}}
\end{array}} \right] \approx 0
\end{numcases}
Since both equation (\ref{eq39}) and (\ref{eq40}) contain $d{\rho _{1d}}$ and  $d{\rho _{2d}}$ , all the higher order terms of  $d{\rho _{1d}}$  and $d{\rho _{2d}}$   can be ignored in these two equations. And equation (\ref{eq39}) and (\ref{eq40}) can be simplified as:

\begin{numcases}{}
{D_3}{\lambda _{11}}d{\rho _{1d}} + {D_3}{\lambda _{21}}d{\rho _{2d}} + {H_{4d}}'\left( {3,3} \right)d{\rho _{3d}}^2 \approx 0\\
{D_3}{\lambda _{11}}d{\rho _{1d}} - {D_3}{\lambda _{21}}d{\rho _{2d}} + {H_{5d}}'\left( {3,3} \right)d{\rho _{3d}}^2 \approx 0
\end{numcases}

So, we have:
\[\left\{ \begin{array}{l}
{D_3}{\lambda _{11}}d{\rho _{1d}} + \frac{{\left( {{H_{4d}}'\left( {3,3} \right) + {H_{5d}}'\left( {3,3} \right)} \right)}}{2}d{\rho _{3d}}^2 \approx 0\\
{D_3}{\lambda _{21}}d{\rho _{1d}} + \frac{{\left( {{H_{4d}}'\left( {3,3} \right) - {H_{5d}}'\left( {3,3} \right)} \right)}}{2}d{\rho _{3d}}^2 \approx 0
\end{array} \right.\]

\begin{numcases}{}
\frac{{d{\rho _{1d}}}}{{d{\rho _{3d}}^2}} \approx \frac{{\left( {{H_{4d}}'\left( {3,3} \right) + {H_{5d}}'\left( {3,3} \right)} \right)}}{{ - 2{D_3}{\lambda _{11}}}}\\
\frac{{d{\rho _{2d}}}}{{d{\rho _{3d}}^2}} \approx \frac{{\left( {{H_{4d}}'\left( {3,3} \right) - {H_{5d}}'\left( {3,3} \right)} \right)}}{{ - 2{D_3}{\lambda _{21}}}} \approx 0
\end{numcases}
That means: $d{\rho _{1d}}$ , $d{\rho _{2d}}$ and $d{\rho _{3d}}^2 ar $e the infinitesimal quantities of the same order.
\begin{gather}\label{eq45}
\left[ {\begin{array}{*{20}{c}}
{d{h_{1d}}}\\
{d{h_{2d}}}\\
{d{h_{3d}}}
\end{array}} \right] \approx \left[ {\begin{array}{*{20}{c}}
{{\lambda _{11}}d{\rho _{1d}} + {\lambda _{21}}d{\rho _{2d}} + {H_{1d}}'\left( {3,3} \right)d{\rho _{3d}}^2}\\
{{\lambda _{11}}d{\rho _{1d}} - {\lambda _{21}}d{\rho _{2d}} + {H_{2d}}'\left( {3,3} \right)d{\rho _{3d}}^2}\\
{{H_{3d}}'\left( {3,3} \right)d{\rho _{3d}}^2}
\end{array}} \right]\end{gather}
From the third equation of (\ref{eq45}), we have the following 2 cases:

(I)	when $d{h_{3d}}/{H_{3d}}'\left( {3,3} \right) > 0$,  $d{\rho _{3d}}$ have 2 real solutions;

(II)and when $d{h_{3d}}/{H_{3d}}'\left( {3,3} \right) < 0$,  $d{\rho _{3d}}$ have 2 imaginary solutions.

So for the point nearby $\left( {{s_{1d}},{s_{2d}},{s_{3d}}} \right)$, in complex space, the Case (I) has 2 real value solutions, and Case (II) has 2 complex value solutions, with non-zero imaginary part. That means, when the optical center passes through CSDC, from equation (\ref{eq45}), we know the 2 P3P solutions change from the real value to complex value with non-zero imaginary part. That means the number of P3P solutions in Case (I) changes by 2 compared with that in Case (II), when the optical center lies outside of the 3 toroids\cite{P3P2019}.
In addition, from equation (\ref{eq45}), we also know that, when the optical center moves on the tangent plane of CSDC, P3P problem could also maintain 2 repeated solution. It is because, in this situation the third element of $\left[ {\begin{array}{*{20}{c}}
{d{h_{1d}}}\\
{d{h_{2d}}}\\
{d{h_{3d}}}
\end{array}} \right]$ equals 0.
By now, we have proved our Proposition 2.

\end{document}